*Article*

# Contingency Analysis of a Grid Connected EV's for Primary Frequency Control of an Industrial Microgrid Using Efficient Control Scheme

**Jayalakshmi N. Sabhahit [1], Sanjana Satish Solanke [1], Vinay Kumar Jadoun [1,*], Hasmat Malik [2], Fausto Pedro García Márquez [3] and Jesús María Pinar-Pérez [4,*]**

[1] Department of Electrical and Electronics, Engineering, Manipal Institute of Technology, Manipal Academy of Higher Education, Manipal 576104, India; jayalakshmi.ns@manipal.edu (J.N.S.); solankesanjana10@gmail.com (S.S.S.)

[2] BEARS, University Town, NUS Campus, Singapore 138602, Singapore; hasmat.malik@gmail.com

[3] Ingenium Research Group, Universidad Castilla-La Mancha, 13071 Ciudad Real, Spain; faustopedro.garcia@uclm.es

[4] Department of Quantitative Methods, CUNEF Universidad, 28040 Madrid, Spain

\* Correspondence: vjadounmnit@gmail.com (V.K.J.); jesusmaria.pinar@cunef.edu (J.M.P.-P.)

**Abstract:** After over a century of internal combustion engines ruling the transport sector, electric vehicles appear to be on the verge of gaining traction due to a slew of advantages, including lower operating costs and lower $CO_2$ emissions. By using the Vehicle-to-Grid (or Grid-to-Vehicle if Electric vehicles (EVs) are utilized as load) approach, EVs can operate as both a load and a source. Primary frequency regulation and congestion management are two essential characteristics of this technology that are added to an industrial microgrid. Industrial Microgrids are made up of different energy sources such as wind farms and PV farms, storage systems, and loads. EVs have gained a lot of interest as a technique for frequency management because of their ability to regulate quickly. Grid reliability depends on this quick reaction. Different contingency, state of charge of the electric vehicles, and a varying number of EVs in an EV fleet are considered in this work, and a proposed control scheme for frequency management is presented. This control scheme enables bidirectional power flow, allowing for primary frequency regulation during the various scenarios that an industrial microgrid may encounter over the course of a 24-h period. The presented controller will provide dependable frequency regulation support to the industrial microgrid during contingencies, as will be demonstrated by simulation results, achieving a more reliable system. However, simulation results will show that by increasing a number of the EVs in a fleet for the Vehicle-to-Grid approach, an industrial microgrid's frequency can be enhanced even further.

**Keywords:** industrial microgrid; wind turbines; PV farm; state of charge; primary frequency control; electric vehicle; vehicle-to-grid

## 1. Introduction

One of the biggest causes of the increasing challenges such as climate change has been fossil fuels. Transportation using conventional fuels is responsible for about 15% of overall world $CO_2$ emissions [1,2]. Renewable power technologies have improved significantly in recent years, as the global energy crisis has become more important [3,4]. More advancement was required with the emergence of power electronics, which necessitated the existence of renewable resources such as wind or PV energies. Rapid innovations in power electronics and batteries have allowed researchers to refocus their attention away from vehicles with internal combustion engine onto electric vehicles (EVs) [5]. Due to their quick regulation ability and superior performance, EVs have been catching the eye as a technology for frequency management [6]. Furthermore, the adoption of distributed energy resources (DERs) has received a lot of attention and interest in the research field to deal with the energy problem and reduce greenhouse gas emissions [7,8]. Parallel to this, the number of industrial microgrids has exploded, owing largely to the consolidation of smaller enterprises into larger ones, as well as the installation of distributed energy supplies.

EVs first appeared in the mid-nineteenth century. With the developments in EV technology, a large demand for EVs arose in the twenty-first century. Many governments, such as the United States and the European Union, have offered incentives to encourage people to buy electric vehicles. The fast adoption of electric vehicles has resulted in the development of a new technology known as the V2G system. The top ten cities in India are among the world's top twenty highly polluted cities. Transportation sources account for 1/3 of pollution, such as particulate matter. As a result, India is a perfect platform for electric vehicles. In 2022, 4.7 GW of electric vehicle storage is expected, according to a source. In Delhi, Chandigarh, and Jaipur, 200 stations of charging will be set up. In India, the Advanced Charging Company made a funding commitment of 1000 crore rupees for infrastructure investment and development.

While the car industry faced another challenging year in 2021, owing to the global semiconductor shortage, global electric vehicle sales more than the doubled in the previous year, hitting 6.6 million units, up from barely 3 million in the 2020. According to the International Energy Agency (IEA), all net gain in worldwide vehicle sales in 2021 could be attributed to the electric vehicles, according to early EV-volumes statistics. China, Europe, and the United States account for nearly 90% of global electric car sales as per the IEA. The fast adoption of electric vehicles has resulted in the development of a new technology known as the V2G system [9].

The idea of Vehicle-to-Grid (V2G) has become major research topic, as it offers valuable ways for industrial microgrids (IMGs) and EVs to interact [10]. Microgrids are employed in modern power systems because they are made up of a variety of DERs such as wind farms and PV farms, and they offer management and control abilities with well-defined boundaries. The bidirectional power transfer and independent nature of microgrids enable them to function like a standalone unit, providing dependability by ensuring supply to essential loads at all instances. Various microgrid research are being carried out, including virtual power plant and autonomous grid [11,12].

There are several different types of microgrids discussed in the literature. One of them is the IMG, which is made up of a few more industrial units connected to renewable energy sources (RES) and energy storage systems (ESS) [13]. The microgrid is generally made up of low-voltage distribution systems that are connected to the power grid via a PCC (point of common coupling) [13]. Controlling the supply–demand balance is a crucial difficulty for a microgrid's successful implementation [14]. The connectivity of renewable sources inside the microgrid, which operates intermittently due to its availability, causes the difference between supply and demand [15]. As a result, EVs could be employed to tackle a frequency fluctuation problem, improving the reliability of the system. Due to the power necessary to keep the batteries charged, electric vehicles will have a detrimental influence mostly on the power grid when they are added to the grid. New modules or existing equipment must be upgraded to counteract this detrimental impact. However, due to the huge expenditure required, these alternatives are not practicable. The cost of updating and installing systems equipment to incorporate the EV inside a fleet with a significant margin inside the present distribution network could exceed 15% [14].

To address this issue, researchers are looking into better charging procedures and energy storage system design so the inclusion of huge numbers of EVs becomes cost-effective and advantageous to both the utility and the consumer [16,17]. However, EVs can benefit microgrids if chargers are built to allow bidirectional flow of power between the grid as well as the ESS. By indicating improved charging techniques, EVs can bring vari-ous advantages to grid. These can reduce power losses while also balancing the load-leveling profile [18]. Interfacing between the grid and EVs requires power electronic equipment that may offer bidirectional power. The bidirectional flow of power is required for V2G technologies to assure high-quality charging/discharging power. This interface among EVs and utilities should perhaps act in response to the utility's charge/discharge signaling. Grid reliability depends on this quick reaction [19].

The challenges of frequency management in microgrids have been addressed [20], including frequency control via charging station operators [21] and droop control method [22,23]. Reference [24] discusses the importance of the power electronics in the industrial microgrid's frequency regulation. The frequency regulation of MGs has been studied by a few researchers. For example, to improve primary frequency stability performance, A. Shokri Gazafroudi et al. [25] employed a droop management method. Models with EVs support are used to evaluate the effective reaction in primary frequency regulation.

An enhanced adaptive droop control technique was provided in references [26,27] because it could better regulate the charging as well as discharging of electric vehicles. EV charging could also be controlled using the real-time droop factor in relation to a primary frequency, EV battery's energy storage, and EV output power control [28]. Iqbal et al. have studied the potential earnings associated to EVs participation in V2G frequency control [29]. In addition to its economic benefit, the associated performance has been investigated because it could be the source of concern for the V2G approach. The "master–slave grid regulation approach" is a technology that relies heavily on intercommunication links and has the disadvantage of lowering system dependability and expanding capacity [30]. Li, Yan et al. suggests that the droop control strategy appears to be the most promising alternative. To develop effective microgrids, droop characteristics were used in power control schemes [31]. The architecture, however, is limited to electronically integrated, rapid, and dispatchable power sources. As a result, the adaptive neuro-fuzzy inference system [32] is introduced. A cutting-edge droop controller also was employed to manage grid frequency employing electric vehicle battery storage [32]. However, this method had the flaw of neglecting numerous critical battery aspects, such as the primary state-of-charge and battery charge or discharge rate. Zhu et al. addresses the issue regarding cutting-edge droop control method [33], in which the researchers used a dual droop synchronized technique to build a V2G strategy and moderate power fluctuations.

Interfacing between the grid and EVs requires power electronic equipment that can offer power flows. Such bi-directional flow of power was required for V2G technologies to assure higher quality charging or discharging electricity. The interface between the EVs and utilities should also consider action in response for the utility's charge and discharge indications. Grid system reliability depends on this fast response [34]. Frequency control via means of charging station operators [35] and the droop-control [36,37] are examples of micro grid frequency control difficulties. References [34,38] discuss the importance of power electronics for industrial micro grid frequency regulation. The primary frequency would also be controlled by the charging station operator, as detailed in [39]. Unlike prior research, our approach takes into account the impact of charge/discharge rates.

As a result of comparing several frequency control strategies through review, we can conclude that the FOPID primary frequency controller is superior to other controllers. This controller has a faster response time to IMG than that of the aforementioned controllers. The goal of this work is to develop a suitable strategy using the FOPID Primary Frequency Controller of an industrial microgrid and contingency analysis of grid connected EV's for frequency regulation. Different EVs charging profiles are analyzed, and a state estimation in % is produced based on the SOC of the charging profile, which defines whether the EV is within charging mode/regulation mode. The following are the primary contributions of this paper:

- A vehicle-to-grid approach is considered for frequency regulation of an IMG by considering proposed primary frequency controller under three different scenarios and each scenario covers three different cases.

- The impacts of the vehicle-to-grid concept for both an IMG and consumers have been investigated for a microgrid.
- The SOC (State-of-charge) is considered to determine the charging current requirements for a EVs charging. Two limiters are also included in the design of the proposed SOC controller to limit the output changes.
- A SE (State Estimation) expressed in % of electric vehicles is also considered to determine the EV in the charging mode or the discharging mode.
- For solving frequency regulation problem, a primary frequency controller is proposed which incorporates fractional order PID controller along with controller for regulating the grid.
- The proposed primary frequency controller is tested for frequency regulation under three different scenarios. The proposed primary frequency controller obtained better results in case 3 than case 2 and case 1.
- The work is validated by comparing different cases for each scenarios study, which results in an industrial microgrid with a variety of distributed energy resources, including solar and wind farms, electric vehicle fleets, residential loads, and a diesel generator.
- The results obtained by the proposed controller on different cases for each scenario are also compared with the latest published work, which validate that proposed controller is effectively perform and provide better results.

The paper is structured as follows: An IMG and its components are outlined in depth in Section 2. The V2G technique for primary frequency control (PFC) is presented in Section 3. Sections 4 and 5 address the system configuration and proposed model. In simulation scenarios and cases described in Section 6, several situations that an IMG may undergo within a 24-h period were simulated, and the frequency variances during these occurrences are analyzed. These scenarios include a PV farm's major reduction in output, a wind farm's tripping, and the start of the synchronous machine. The results are presented in Section 7, which shows that using a FOPID primary frequency controller as well as by increasing the number of EVs in a fleet frequency will be improved to an even better level. The SOC of the different cars in the fleet is divided into the five types of car profile. Each car profile contains several EVs. According to the three different cases, we have added cars in each profile. A case study is designed and investigated using the MATLAB/SIMULINK software. Finally, Section 8 provides the conclusions of this study.

## 2. Industrial Microgrid and Its Subsystems

EVs' ability of quick regulation and superior performance has been catching the eye as a technology for frequency management. Furthermore, the adoption of distributed energy has garnered a lot of attention in the research field to deal with the energy problem and reduce greenhouse gas emissions. Parallel to this, the number of IMGs has exploded, owing largely to the consolidation of smaller enterprises into larger ones, as well as the installation of distributed energy supplies.

A diagram illustrating the industrial microgrid is shown in Figure 1. There are several different types of microgrids. Among them is the IMG, which is made up of several industrial units that are connected to renewable energy source (RES) and energy storage systems. Low-voltage distribution network connects the microgrid to the electrical grid via a PCC. When compared with a traditional grid, the capacity of DERs connected to an IMG is comparatively low. As a result, they are known as micro-sources. Photovoltaic systems, wind energy, electric vehicles, and diesel generator are all examples of micro-sources which are shown in Figure 1.

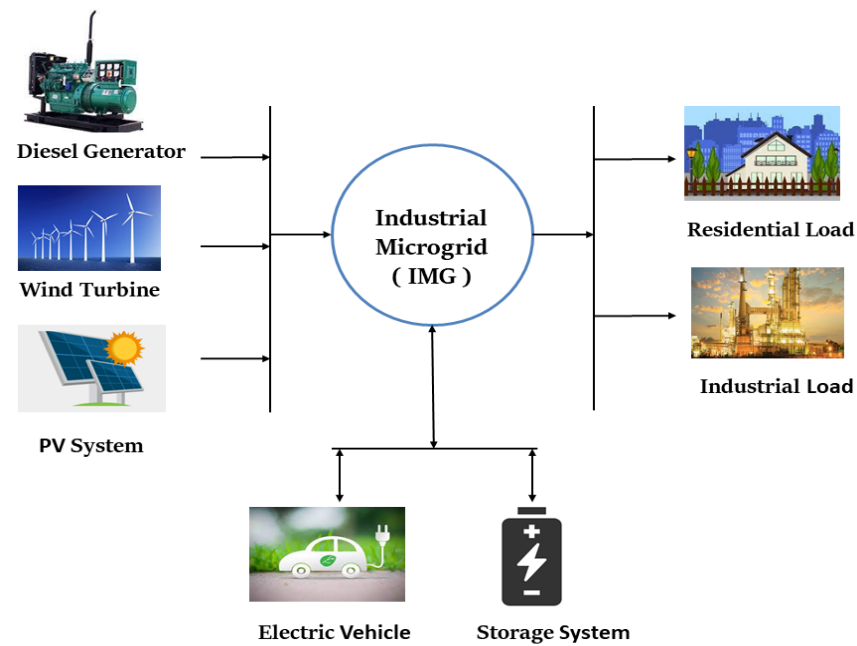

**Figure 1.** Conceptual view of an industrial microgrid.

V2G technology has advantageous to both IMGs and customers. Owners of electric vehicles will not only charge their own vehicles at a lower rate, but they will also sell power to IMG throughout peak hours. Owing to the use of V2G technologies, there is no additional cost of operation for building additional plants of power to offer additional electricity during peak demand, which is helpful to power grids.

### 2.1. Microgrid Subsystems

#### 2.1.1. Diesel Generator

The challenges connected with the unstable and unexpected existence of RESs can be solved by using a diesel generator as a standby or backup power supply. The amount of electricity generated by a diesel generator, diesel engine governor parameters, and excitation system parameters are all important factors. The purpose of a diesel generator is to generate electricity by using a diesel engine governor to drive synchronous generator. Many components make up a diesel generator, including the diesel engine governor's, synchronous generator, the starter winding and the field winding parameters, voltage regulation, ignition systems, etc.

The alternating current signal has a frequency of 50 Hz, and the engine's speed is constant due to its regulator time system. The projected primary power generator has a 15 MW capacity. It has the capacity to adjust voltage and frequency, as well as improve system reliability.

#### 2.1.2. Solar PV Farm

The amount of energy which is produced by a photovoltaic farm was determined by the covered area of panels as well as the amount of the irradiance generated. The output current of a solar cell is directly related to the amount of light falling on it. These input parameters for determining the current, power, and voltage as an output factor are irradiance and temperature. Due to its low cost and lack of machine-driven parts, PV energy is gaining more and more attention among all RESs.

PV systems also have the advantage of being simpler to install than some other types of RESs. Using advanced power switching converters, most of RESs are connected to microgrid/utility grid. The Incremental Conductance Algorithm (INC) was utilized in this 8 MW PV system because it can follow rapidly rising and dropping irradiance situations with more accuracy. The number of solar cells added to a system determines the PV cell's overall output capacity. An inverter is used to convert the DC power on PV systems side to the AC power mostly on microgrid side when a PV systems output is linked to the microgrid.

The current generated by a solar panel (I) is equivalent to the current generated by current source, minus the current that flows over a diode, and minus the current that flows into shunt resistors, as shown in an equivalent circuit of the solar cell in Figure 2.

$$I_{PV} = I_L - I_{OS} - I_{SH} \tag{1}$$

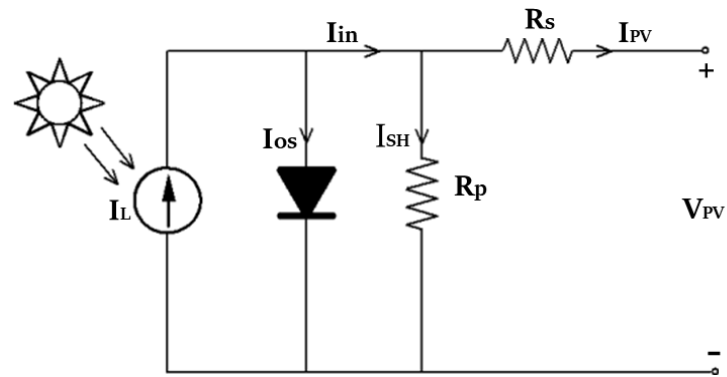

**Figure 2.** An equivalent circuit of a solar cell.

Voltage (V) across these elements controls the current that flows through them.

$$V = V_{PV} + I_{PV} \cdot R_S \tag{2}$$

The current redirected through a diode, according to the Shockley diode equation, is:

$$I_{OS} = I_O \left\{ \exp\left[ \frac{V}{\xi V_T} \right] - 1 \right\} \tag{3}$$

The current redirected through shunt resistor ($R_{SH}$), according to Ohm's law, is:

$$I_{SH} = \frac{V}{R_{SH}} \tag{4}$$

$$I = I_L - I_O \left\{ \exp\left[ \frac{V_{PV} + I_{PV} \cdot R_S}{\xi V_T} \right] - 1 \right\} - \left[ \frac{V_{PV} + I_{PV} \cdot R_S}{R_{SH}} \right] \tag{5}$$

Figure 2 shows how the PV cell's output current is calculated. Photo-generated current is defined by $I_L$, where $I_{OS}$ is the saturation current of diode, $I$ is the output current, $I_{SH}$ is the shunt current, $I_{PV}$ is the current across PV cells, $\xi$ is the diode ideality factor, $V_T$ represents the thermal voltage, $I_0$ is diode current and $R_S$ is series resistance that represents losses as from interconnections, $R_{SH}$ is the shunt resistor, and $V_{PV}$ is the voltage across PV cells. The number of solar cells added to a system determines the PV cell's overall output capacity. An inverter is used to convert the DC power of the PV system to AC power of a microgrid side when a PV farm output is linked to the MG [40].

### 2.1.3. Wind Farm

A wind turbine using the doubly fed induction generator comprised of the wound rotor induction generator is simulated in this work. The stator is coupled to a grid directly, while a rotor was fed at a variable frequency by the converter. The power generated by wind turbines is being regarded as a variable renewable source since it is dependent upon a wind profile, which changes over time. Equation (6) [41,42] are used to determine the output power of Wind Turbine.

$$P = \frac{1}{2} \rho A V^3 \tag{6}$$

where $A$ is the swept area of the blades, $\rho$ is the air density, and $V$ is the velocity of the wind, respectively. Due to the time-variant air direction and speed, the resultant output of the wind turbine varies as a natural resource. In terms of performance of the distributed generator and the EV's controller, the fundamental properties of wind turbines also have some impact on the frequency of the microgrid.

### 2.1.4. Load

The load of the system is made up of various components, including residential load and, also, an asynchronous machine (ACM), which are then used to simulate the influence of the industrial load (inductive) on the microgrid. A home load's power factor is followed by the consumption profile. A square relationship between rotor speed and the mechanical torque controls the ACM. Table 1 collects the load parameters used in this study.

**Table 1.** Load parameters.

| Parameters | Value |
|---|---|
| Nominal Power | 10 MW |
| Power Factor | 0.95 |
| Asynchronous Machine (Squirrel-cage) | 0.16 MVA |
| Time Step | 60 |

### 2.1.5. Electric Vehicle

This study focuses on the role of EVs in a microgrid's frequency modulation. Internal energy loss and the battery type technology are not considered in this work. These assumptions have no bearing on the study's goals. EV has created numerous aggregation models to investigate the microgrid's frequency regulation.

The storage system for electric vehicles is a power source with bi-directional flow of power feature [43]. As a result, under the guidance of an efficient controller, EVs had a tremendous potential to deliver power to an IMG to stabilize the frequency of a microgrid throughout peak time hours.

EV's Simulink model is presented in Figure 3. When plugged in with industrial microgrid, the EVs perform two primary functions: EVs behave as loads and absorb charge. Second, EVs could be employed as a storage unit to deliver power to a load, satisfying the system's demand and providing ancillary frequency control services. The widely used first order EV transfer function is described as follows:

$$PEV = \frac{Kev}{Tevs + 1} \tag{7}$$

where $Kev$ and $Tevs$ are the EV battery's gain and time constant respectively. The parameters $Kev$ and $Tev$ values are considered as 0.333 and 1, respectively, from theoretical calculations using formulas described in reference [44]. The primary purpose of this method is to maximize the use of renewable energy supplies while reducing reliance on fossil fuel plants for a clean and green environment [45–47].

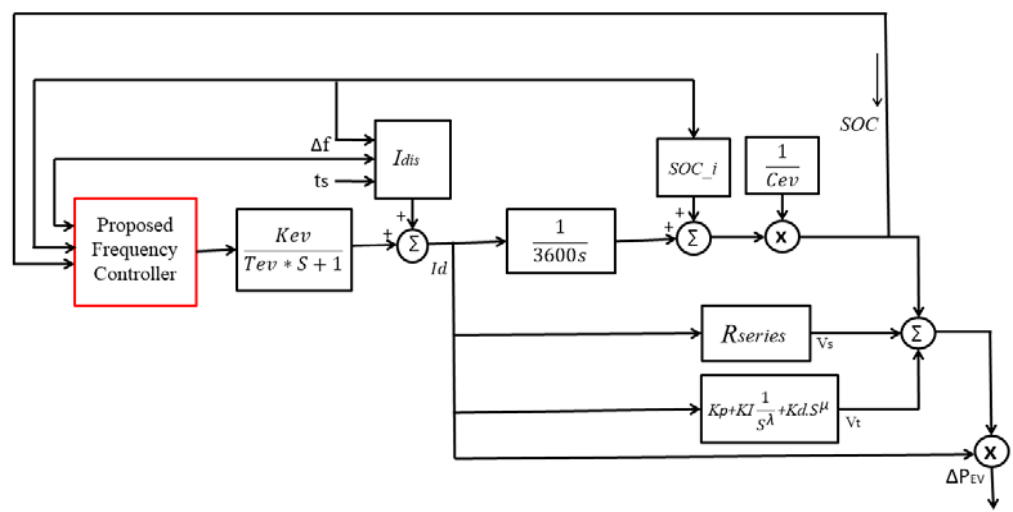

**Figure 3.** EV model.

## 3. V2G Primary Frequency Control Strategy

The V2G model is advantageous to EV owners and addresses the IMG's stability difficulties. The core principle is to use the batteries of EVs as a source of intermediate energy. EVs electricity is fed into a grid at peak times. Excess power is used to charge many EVs during off-peak periods due to low demand of load.

This system is extremely valuable to both IMG as well as consumers. Owners of EVs will not only charge their own vehicles at a lower rate, but they will also sell electricity to IMG at peak periods. Due to the use of V2G system, there are no substantial additional costs for building extra power plants to give extra power during peak demand times, which is beneficial to electrical grids [48,49]. This system requires the installation of a charging station, which is a location where electric vehicles may be plugged in to charge their batteries. EVs will be supplied with power by IMG via charging stations as needed. During periods of heavy demand of load, power should be returned to the grid via the station of charging. Figure 4 depicts the V2G framework. The system works as follows:

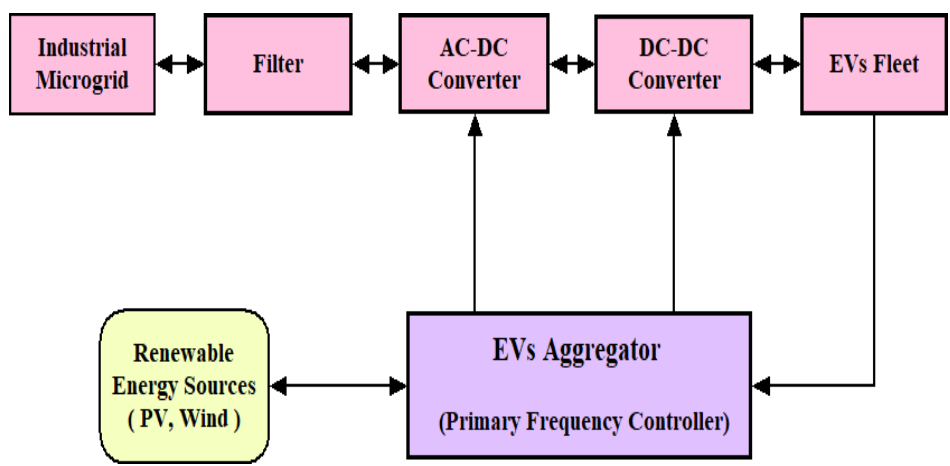

**Figure 4.** Block diagram of V2G system for an IMG.

For G2V Mode: $P_T > 0$, EV works in charging mode.

For V2G Mode: $P_T < 0$, EV works in discharging mode.

$P_T$ is the total power. When $P_T < 0$, EV experienced rapid discharge, this resulted in a frequency fluctuation. Consequently, the objective function for these two approaches was built on the charging or the discharging techniques of EVs. The information is provided from the IMG to the EV aggregator to disconnect an EV inside a discharging mode.

## 4. System Configuration

The interface of V2G and the electricity grid is depicted in Figure 5. The proposed microgrid comprises of the base power generator such as a diesel generator, sources of renewable energy, such as solar and wind energy, as well as a V2G system, which are all connected via Bus 1, Bus 2, and Bus 3. The base model includes five electric vehicles profiles and residential load. The proposed work additionally includes a control to enable or disable V2G mode. This charging technique is bidirectional, which means it may both supply and consume power in amongst grid and the charging station.

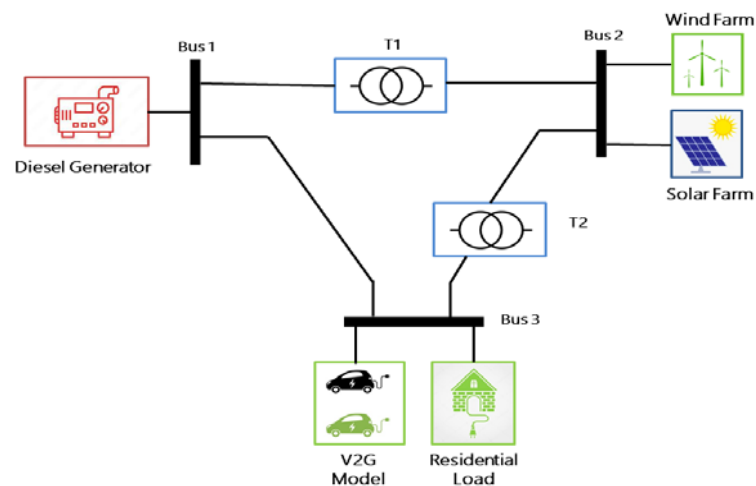

**Figure 5.** Design of the system.

Table 2 presents the simulations parameters used. The PV farm's nominal power is 8 MW, while tripping of the solar farm is being simulated around 12 p.m., indicating cloudy weather daytime. A wind farm has a capacity of 4.5 MW with a normal wind speed of the 13.5 m/s. When the wind speed reaches 15 m/s, a wind farm would disconnect out from system and reconnect when the wind speed reaches its normal speed.

**Table 2.** Simulation parameters.

| Technical Parameters | Value |
| --- | --- |
| Wind Energy (MW) | 4.5 |
| Speed of the wind (m/s) | 13.5 |
| Max speed of the wind (m/s) | 15 |
| PV Farm Power (MW) | 8 |
| Diesel Generator (MW) | 15 |
| Nominal Frequency (Hz) | 50 |
| Dead Band | 0.001 |
| SOC (min) | 0.2 |
| SOC (max) | 0.8 |
| Bus 1—Base Voltage (KV) | 25 |
| Bus 2—Base Voltage (KV) | 25 |
| Bus 3—Base Voltage (KV) | 1 |
| V2G (MW) | 4 |

The nominal power of the diesel generating system is 15 MW, with the potential of 25 kV as well as a frequency of 50 Hz. When there was a disruption in a microgrid, the diesel generator's main duty is to balance demand of power. Since it is linked to the electricity grid, the diesel generator balances power supply as well as load demand, and any changes in grid frequency are seen through a rotor speed.

## 5. Proposed Primary Frequency Controller

The fractional order PID controller is a modification on the standard PID controller. The fundamental advantage of the fractional-order control is that it allows the control system to have more customizable time–frequency responses, allowing for better and reliable performance. FOPID controller has received a lot of interest over the last few years, from both the academics and from industry [50].

In this controller, we have incorporated a FOPID controller along with a controller for grid regulation. Figure 6 depicts the proposed controller for primary frequency regulation; the main goal of this controller is to sustain the industrial microgrids during the failure of the any source as well as the addition of a large load to industrial microgrids. Frequency analysis is required for effective regulation of grid. A full block design of a proposed controller is shown in Figure 6.

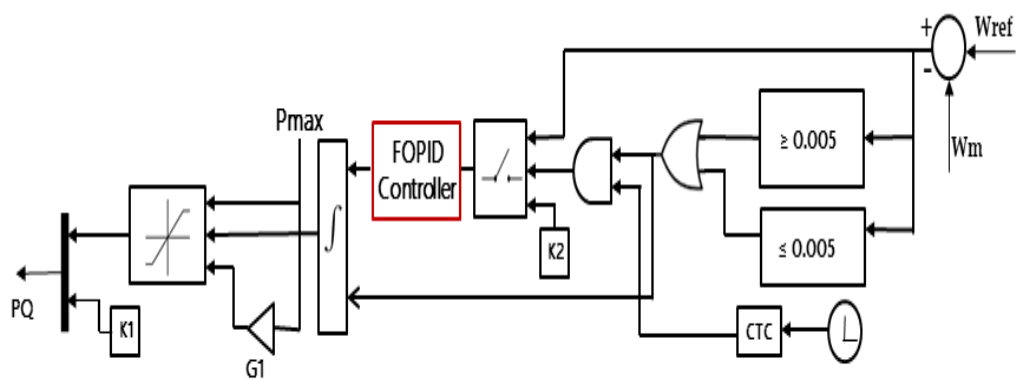

**Figure 6.** Proposed primary frequency controller.

The frequency could be managed even better with raising the number of EV, according to the simulation. During an under-frequency scenario, the controller supervises the functioning of the industrial grid by regulation mode or the discharging mode when any one of the sources is tripped and any large industrial load is put into service.

Frequency measurement, often known as the grid frequency, is required for effective regulation of grid. We can determine system frequency variances by comparing grid frequency (Wm) to a reference frequency (Wref), as shown in the proposed model an input of both the frequencies is given to the FOPID controller.

Logic gates (AND and OR) will decide whether it is in the charging mode/regulation mode. The main decision will take place here with the help of FOPID controller. Gain parameter (G1) decides the rate of change of voltage for determining the mode of operation. The CTC (Compare to constant) block is used according to which clock input is given. This block is used because the rate of change with respect to time interval is needed. The threshold (K2) with a value of 0.5 is used in the controller design as shown in proposed model. This controller will pass the input of the first stage to the second stage when the input crosses the threshold. A derivative was created to reduce the number of sudden variations in frequency. All the values of simulation parameters are shown in Table 3.

**Table 3.** Simulation parameters for the proposed model.

| Parameters | Value |
|---|---|
| Proportional gain | 1 |
| Integral gain | 0.02 |
| Threshold | 0.5 |
| Derivative gain | 0.01 |
| Gain (G1) | −1 |
| Constant (K1) | 0.01 |

## 6. Scenarios and Cases

Three alternative scenarios are used in this work: First, there is Scenario 1, in which there is reduced PV farm power generation. In Scenario 2, tripping of the wind farm is considered, and Scenario 3 is proposed for asynchronous load starting. Each scenario covers three different cases. A renewable energy source has been incorporated into the Scenario 1 and scenario 2. In order to validate the simulation results of a proposed model, these three scenarios are compared on all the three cases of the system. Different eventualities well over course of the day are simulated in this work, and the V2G approach is validated for the frequency regulation.

The following scenarios are considered in this work:

- Scenario 1: Reduced PV farm power generation.
- Scenario 2: Tripping of the wind farm.
- Scenario 3: Asynchronous load starting.

The details of all three cases considered for every scenario are given below:

- Case 1: V2G mode is deactivated
- Case 2: V2G mode is activated (with 100 EVs)
- Case 3: V2G mode is activated (with 200 EVs)

The proposed model is implemented and simulated in a MATLAB (R2020a version)/SIMULINK environment on Intel i5 processor with 8 GB RAM.

## 7. Result Analysis

Different eventualities well over course of the day were simulated in this work, and the V2G approach was validated for the frequency regulation. These eventualities include three cases which are as V2G mode is deactivated, V2G mode is activated (100 EVs), and V2G mode is activated (200 EVs).

### 7.1. Result Analysis for Scenario 1

The simulations for reduced PV farm power generation are performed in the first instance. PV farm output drops substantially on cloudy days, generating only up to 20% of its rated capacity. Figure 7a, b depicts an output power graph of an occurrence. Figure 7c, d depict the micro-grid's total load and generation. A solar farm will generate power on roughly 25,000 s (07 am) in Figure 7a, b, and then it continues to increase. However, around 43,200 s, or exactly 12 o'clock, there is a sharp drop in solar output, indicating a fall in power to 20 percent of the nominal power.

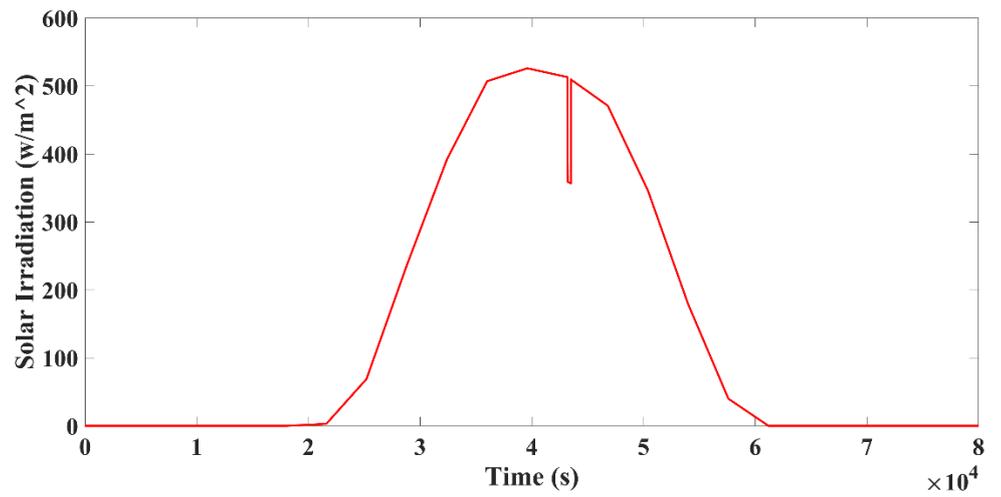

(a)

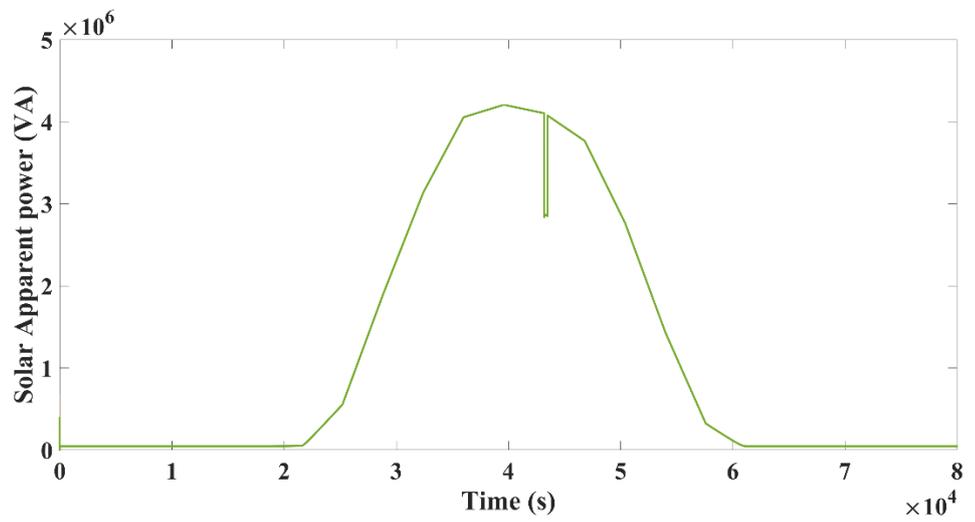

(b)

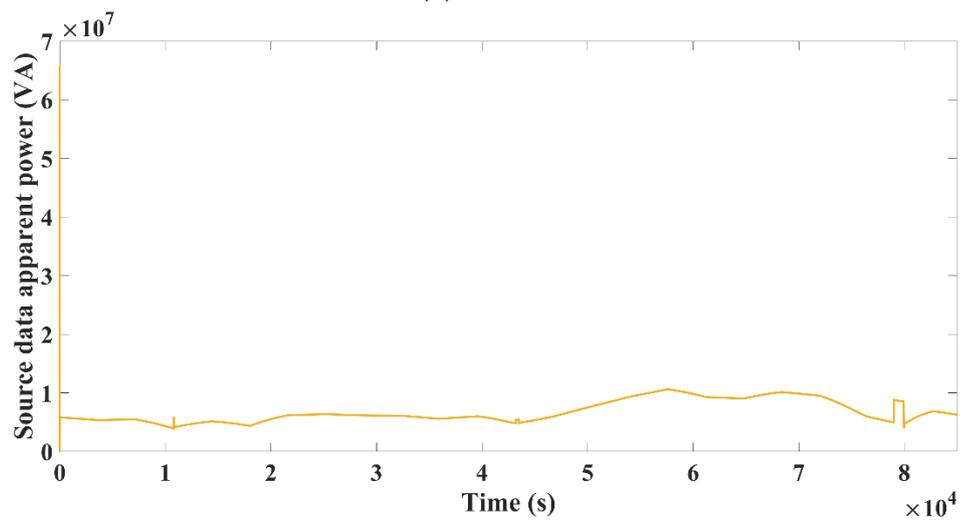

(c)

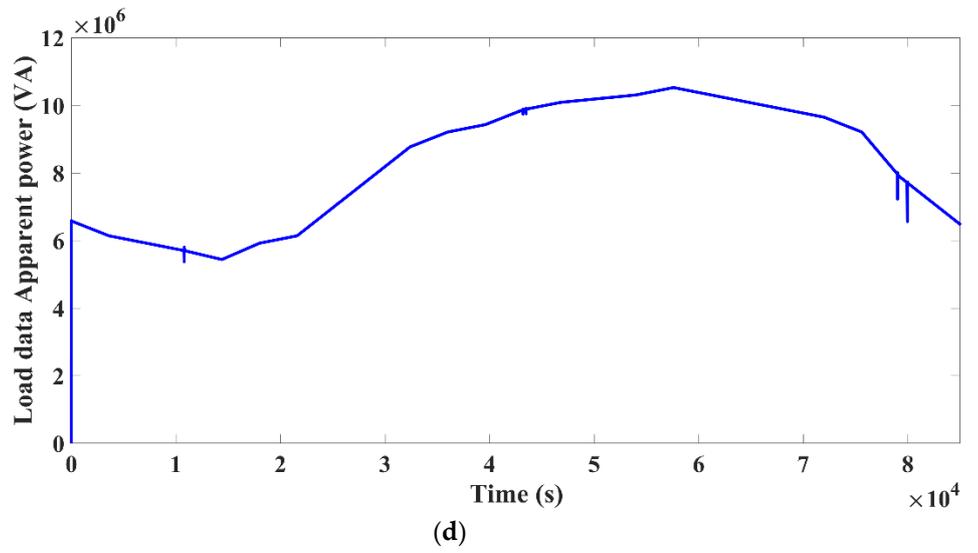

(**d**)

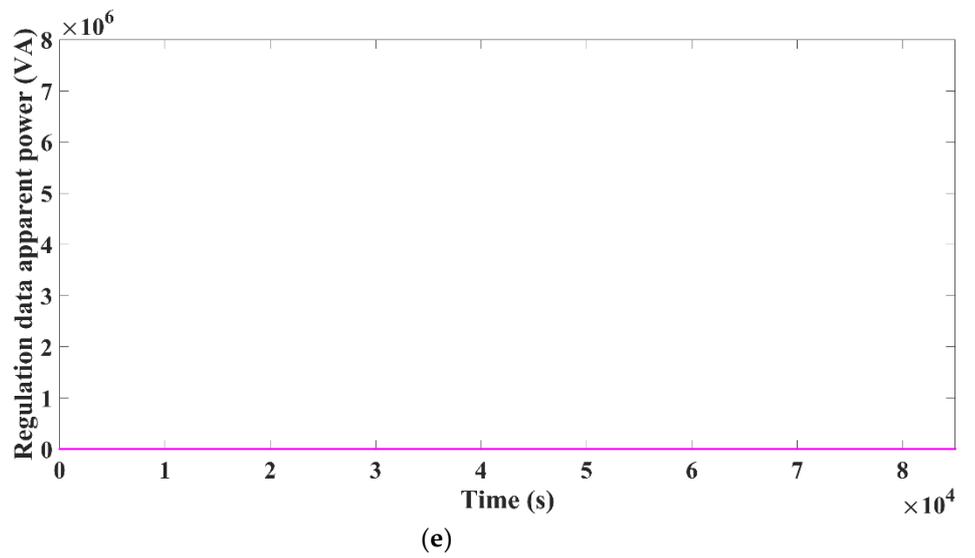

(**e**)

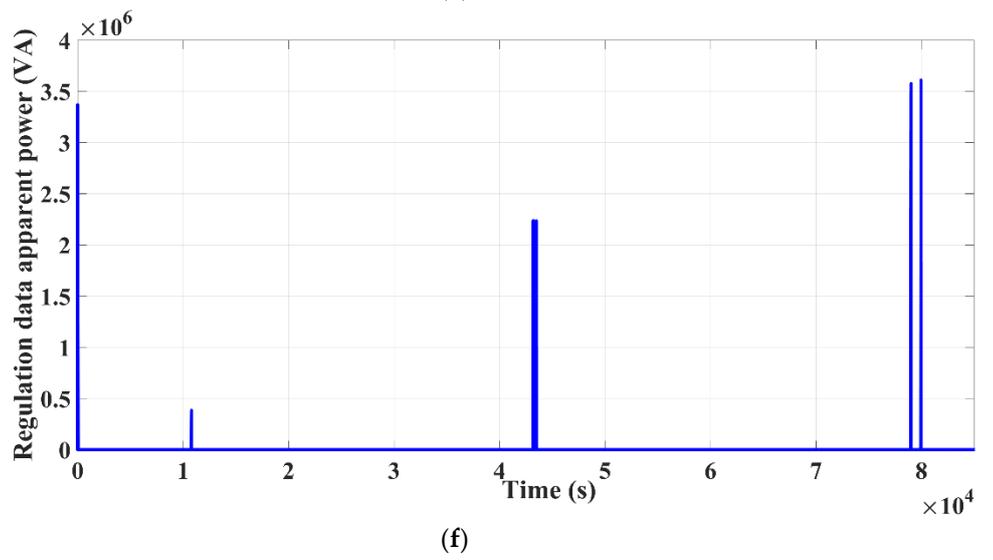

(**f**)

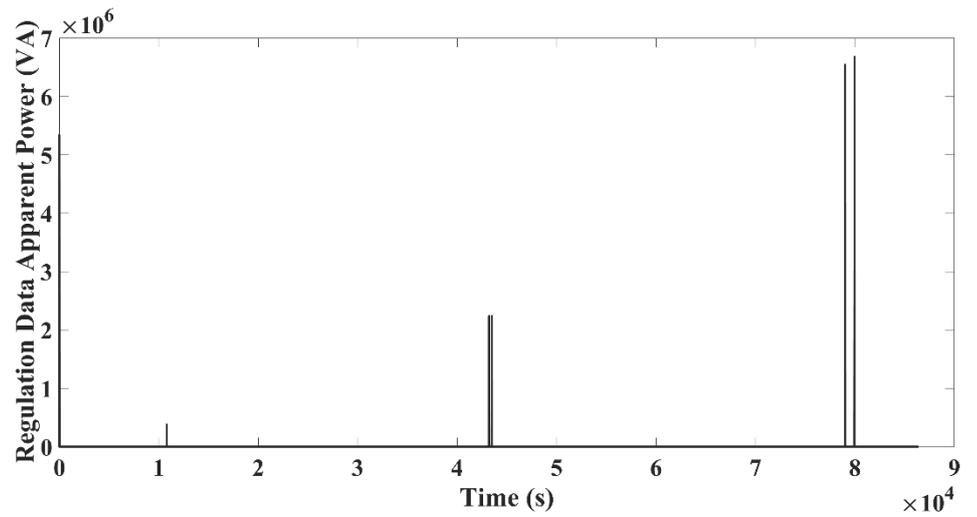

(**g**)

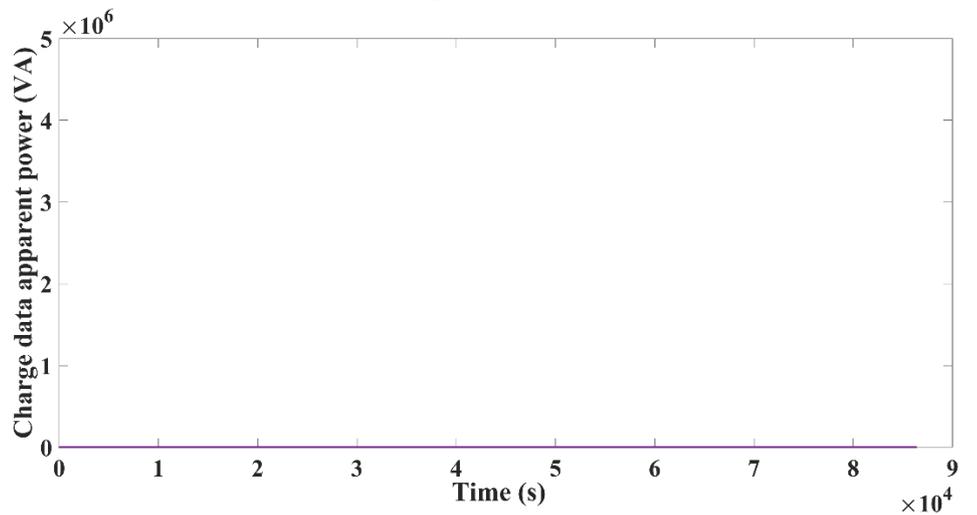

(**h**)

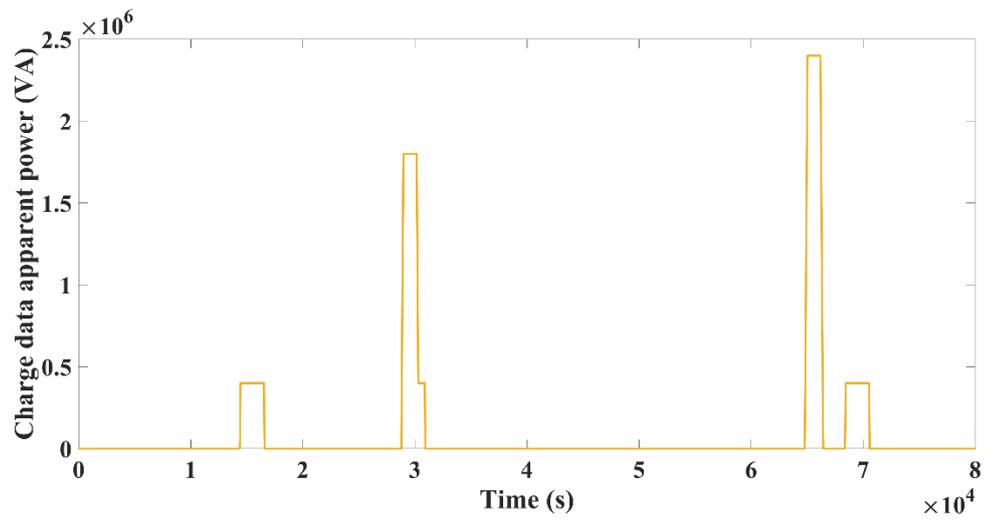

(**i**)

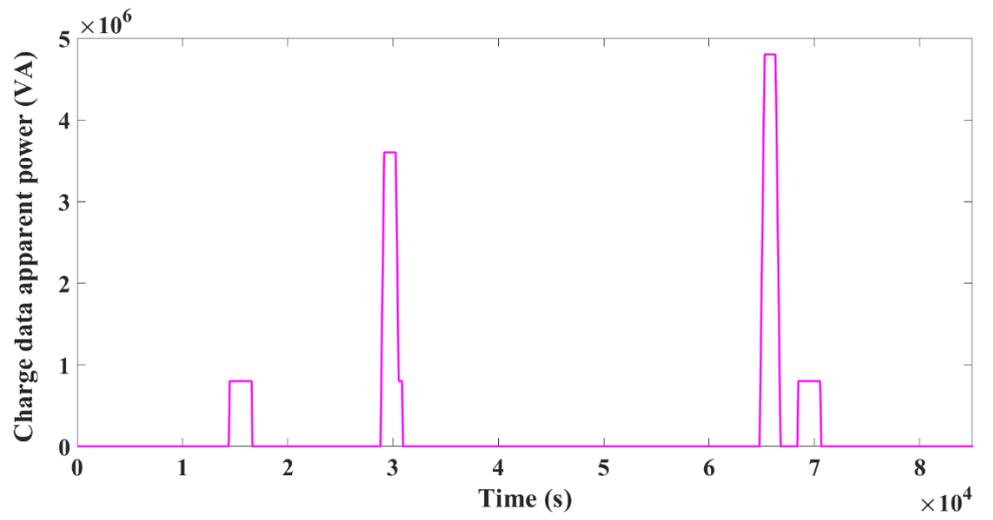

(j)

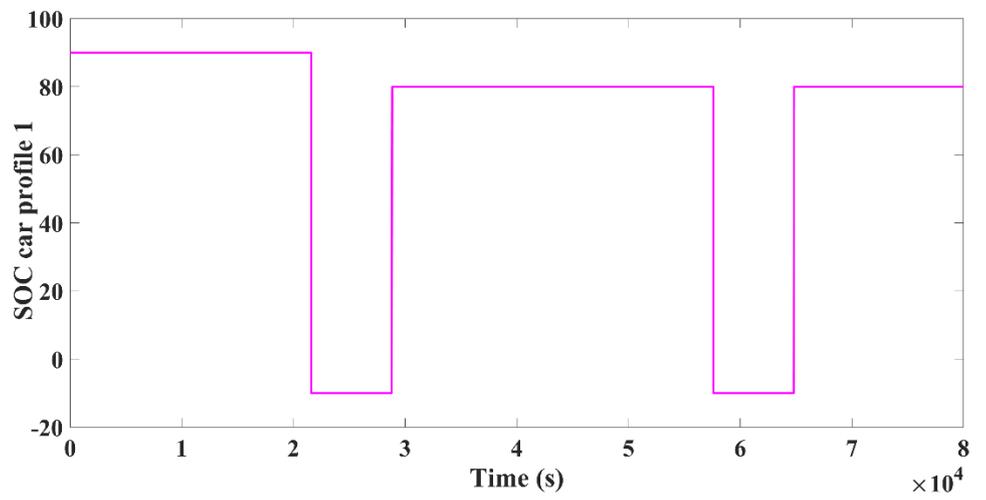

(k)

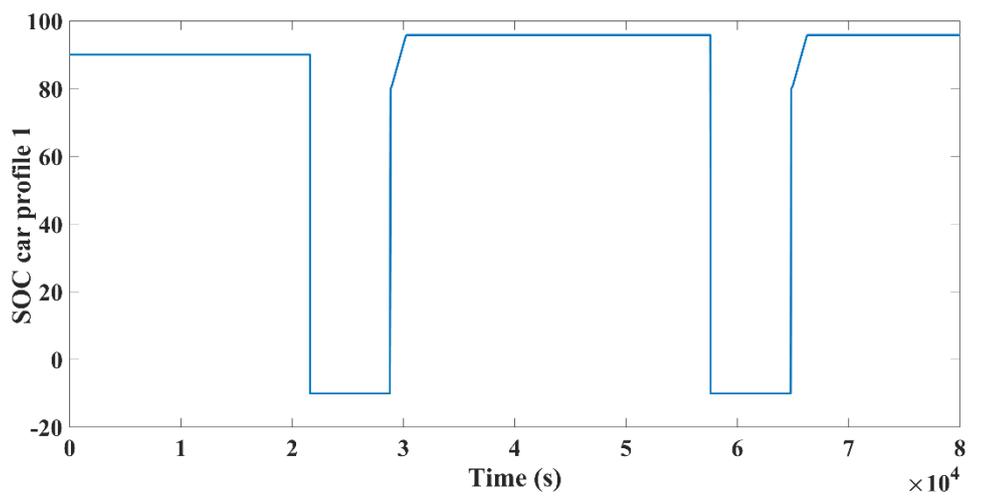

(l)

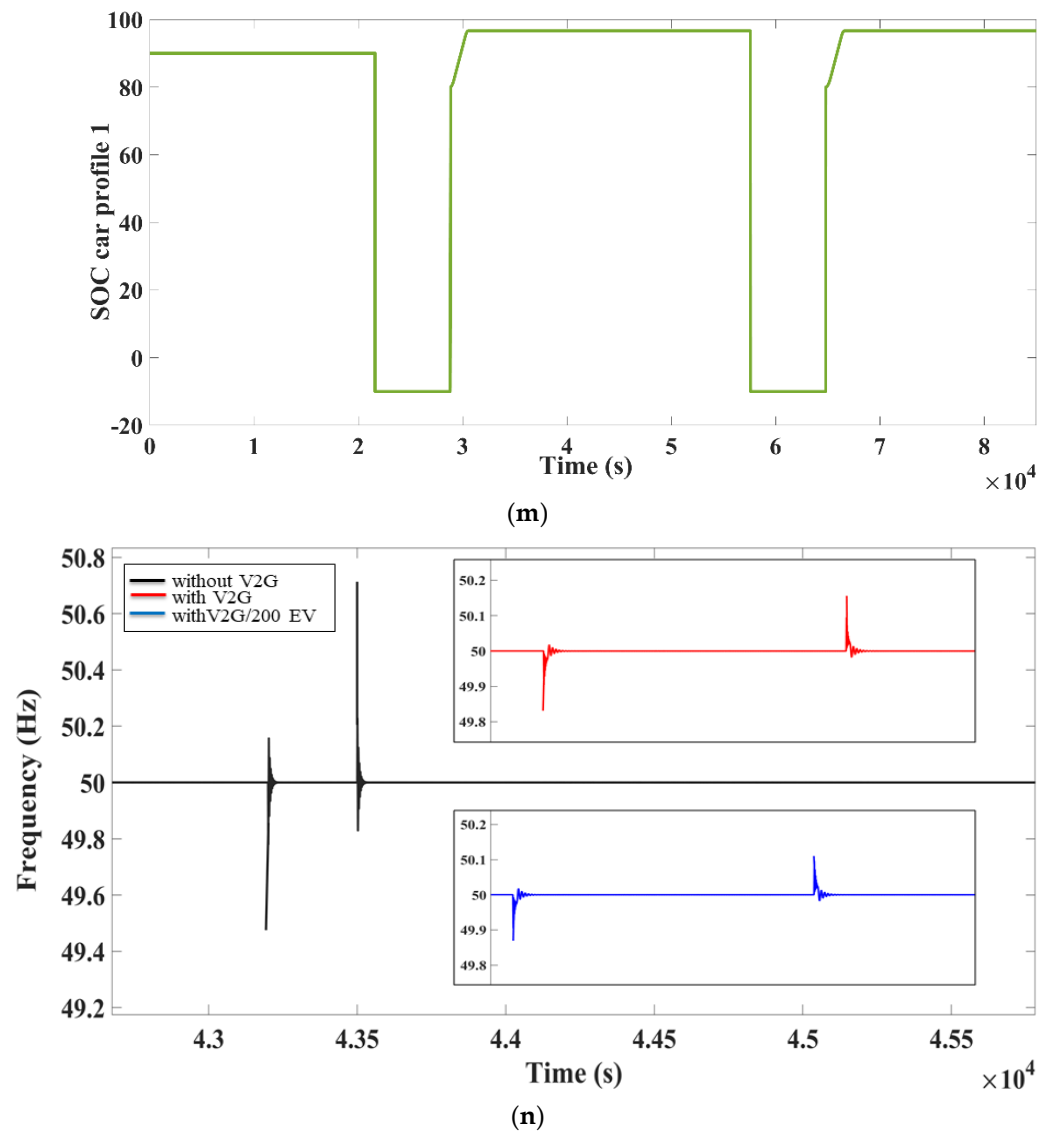

**Figure 7.** Reduced PV Farm generation: (**a**) solar irradiation, (**b**) apparent power, (**c**) total power generation, (**d**) load data for case 1, (**e**) regulation data for case 1, (**f**) regulation data for case 2, (**g**) regulation data for case 3, (**h**) charge data for case 1, (**i**) charge data for case 2, (**j**) charge data for case 3, (**k**) SOC car profile for case 1, (**l**) SOC car profile 1 for case 2, (**m**) SOC car profile for case 3, and (**n**) frequency regulation in 3 different cases.

This severe dip lasted for about five minutes, following which the solar power began to generate power at full capacity at 43,500 s. When the generation was superior to the load, the V2G operates in charging mode, as shown in Figure 7h–j. In the same way, if the load exceeds the generation, the EV's will enter regulation mode, as shown in Figure 7e–g. The system includes the different vehicle profiles, and Figure 7k–m depicts the SOC for car profile 1 in three cases. Under three different scenarios, each disturbance inside an IMG is tested for the frequency regulation.

Under three different circumstances, the disturbance in the IMG is tested as frequency regulation. The V2G mode disables the principal frequency in the first case, as shown in Figure 7n, which is highlighted in black. In second instance, when V2G method was enabled, as indicated in Figure 7n, the IMG frequency quickly stabilized and changed, as shown in red color. In the same way as the third instance, when a number of EVs in a fleet rise, an industrial microgrid frequency rises to even a greater level, as seen in Figure 7n, by the blue color. Table 4 presents the comparison of frequency regulation for three different cases. The result obtained by proposed controller has better frequency regulation compared with reference method.

**Table 4.** Comparison of results for three separate cases in scenario 1 using the proposed controller compared to those of Reference.

| Mode | V2G Off | | V2G On (100 EVs) | | V2G On (200 EVs) | |
|---|---|---|---|---|---|---|
| | Ref. [51] | Proposed Controller | Ref. [51] | Proposed Controller | Ref. [51] | Proposed Controller |
| Frequency (Min Value) | 49.42 | 49.44 | 49.79 | 49.82 | 49.81 | 49.86 |
| Frequency (Max Value) | 50.75 | 50.72 | 50.20 | 50.17 | 50.16 | 50.12 |

### 7.2. Result Analysis for Scenario 2

Wind farm simulations have been carried out in this case. When the wind speed surpasses 15 m/s, which is maximum speed restriction, the wind farm trips. This is to prevent damage to the wind turbine due to the increased wind speeds. When the speed of wind surpasses its normal speed of 13.5 m/s, the wind farm reconnects to the grid. Figure 8a depicts typical profile of wind during a 24-h period. The wind farm tripped after 79,200 s, which equates to 10 p.m. The wind speed will be more than that of the top set point of the 15 m/s at this time, and it returns to the normal output after a period of time when the speed of the wind returns to the nominal limits. Figure 8c–e depicts the state-of-charge of the car profile 2 for the three different cases.

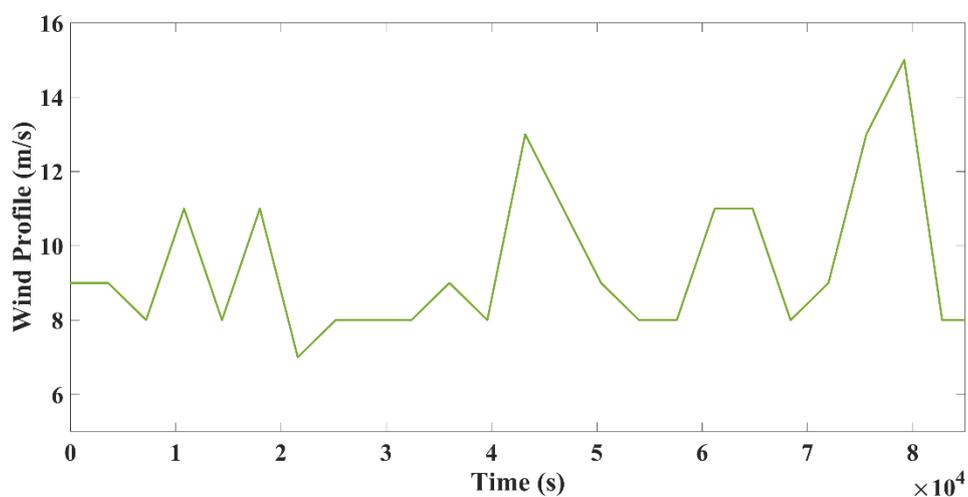

**(a)**

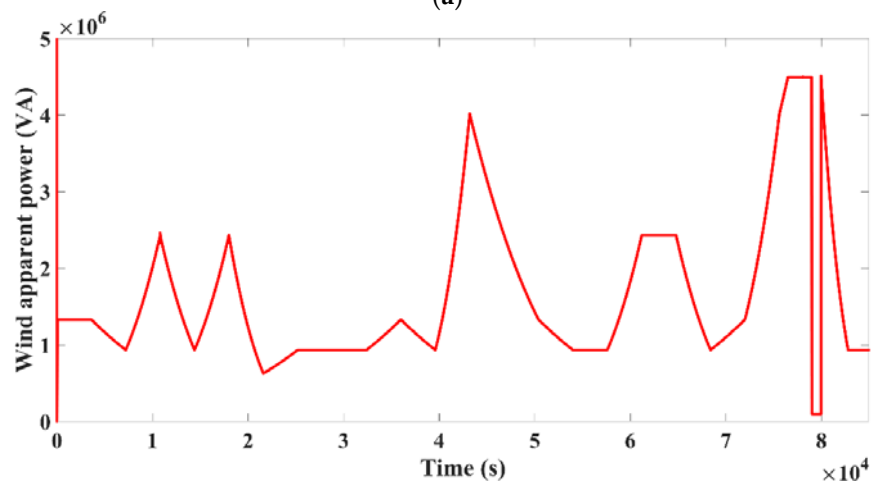

(**b**)

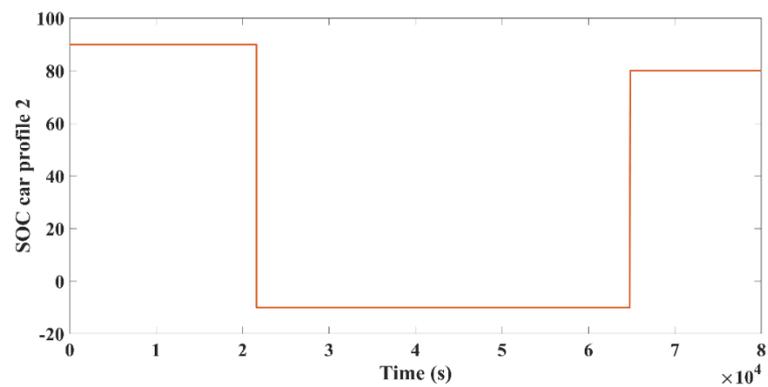

(**c**)

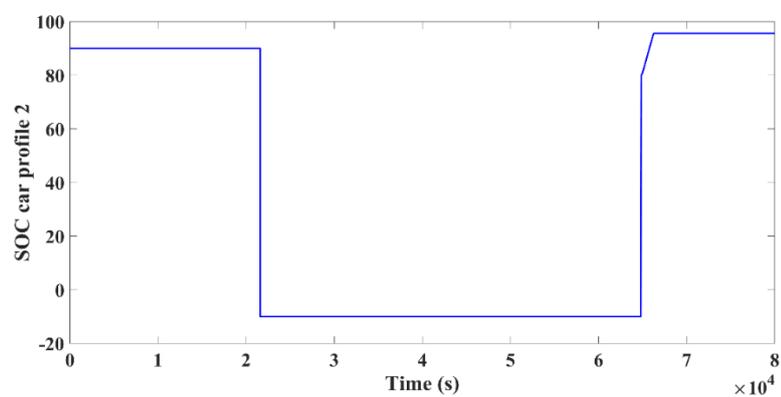

(**d**)

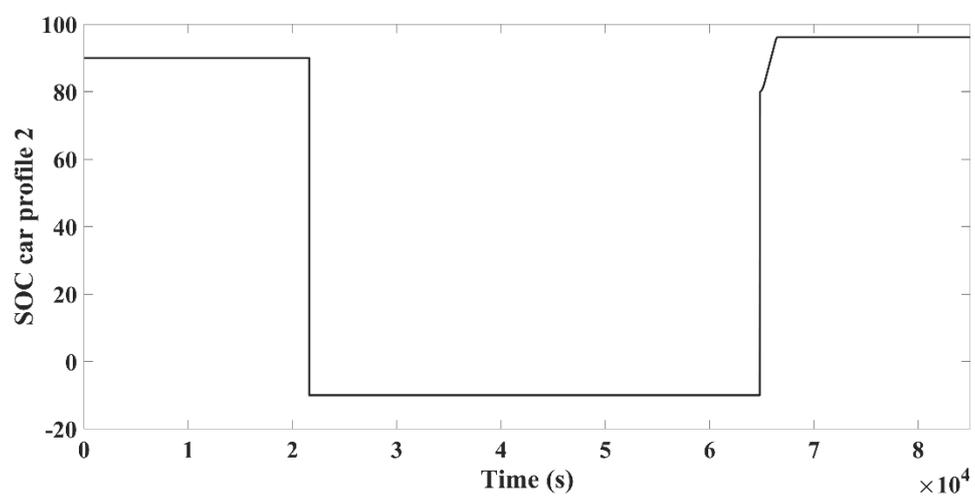

(**e**)

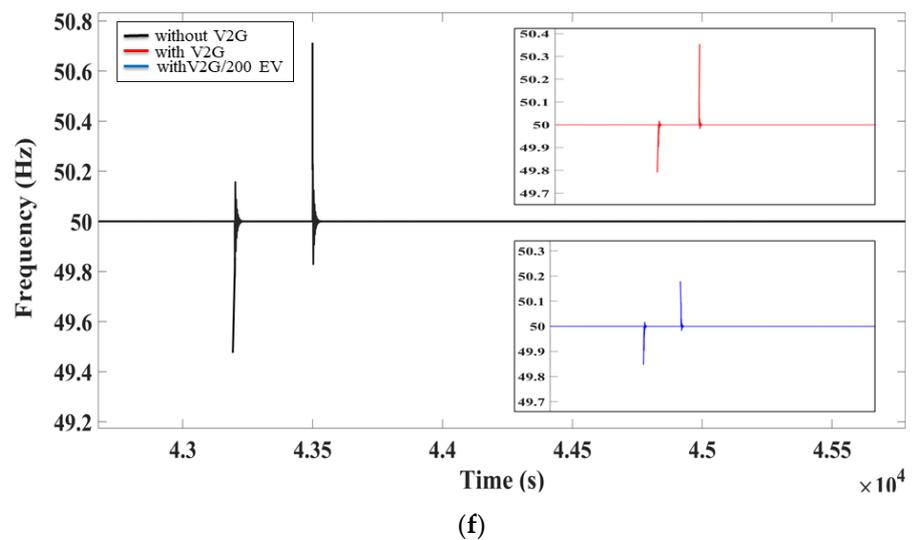

**(f)**

**Figure 8.** Tripping of wind farm: (**a**) wind profile, (**b**) wind apparent power, (**c**) SOC car profile for case 1, (**d**) SOC car profile 1 for case 2, (**e**) SOC car profile for case 3, and (**f**) frequency regulation for 3 different cases.

Figure 8b illustrates the scenario during a tripping of the wind turbine when V2G mode is inactive. It is worth noting that the dip is more pronounced than in case of the solar farm as the solar production is zero at night. This drop in frequency was enough to consider the grid was unstable, as well as that the load should be reduced to compensate for the grid's negative impact. When V2G function is activated, the EVs inside the microgrid help to regulate frequency and increase the microgrid's stability when a disturbance arises in source trip. The frequency of a grid was improved to even more acceptable bounds by raising the number of EVs in fleets from 100 to 200, as presented in Figure 8f. Furthermore, whenever the number of EVs rises, frequency stabilizes both in under and over frequency zones. Table 5 presents a summary for the deviations of frequency for all the cases. The result obtained by proposed controller has better frequency regulation compared with reference method.

**Table 5.** Comparison of results for three separate cases in scenario 2 using the proposed controller compared to those of Reference.

| Mod | V2G Off | | V2G On (100 EVs) | | V2G On (200 EVs) | |
|---|---|---|---|---|---|---|
| | Ref. [51] | Proposed Controller | Ref. [51] | Proposed Controller | Ref. [51] | Proposed Controller |
| Frequency (Min Value) | 49.36 | 49.39 | 49.70 | 49.78 | 49.8 | 50.21 |
| Frequency (Max Value) | 50.90 | 50.80 | 50.35 | 50.32 | 50.21 | 50.19 |

### 7.3. Result Analysis for Scenario 3

The impact of the load on grid frequency was considered in this case. In past circumstances, V2G has been shown to assist in reducing frequency variations to tolerable levels when renewables are tripped from the grid. There is an induction machine in this case, which starts at midday. If the normal power of the asynchronous motor is high, it is known that as soon as the induction machine (IM) begins to run and it destabilizes a grid frequency, which it cannot recover to a long time. Starting current was on the order of six to eight times, which can result in a system with a sudden inrush current. Figure 9a depicts the induction motor starting at midday (43,200 s).

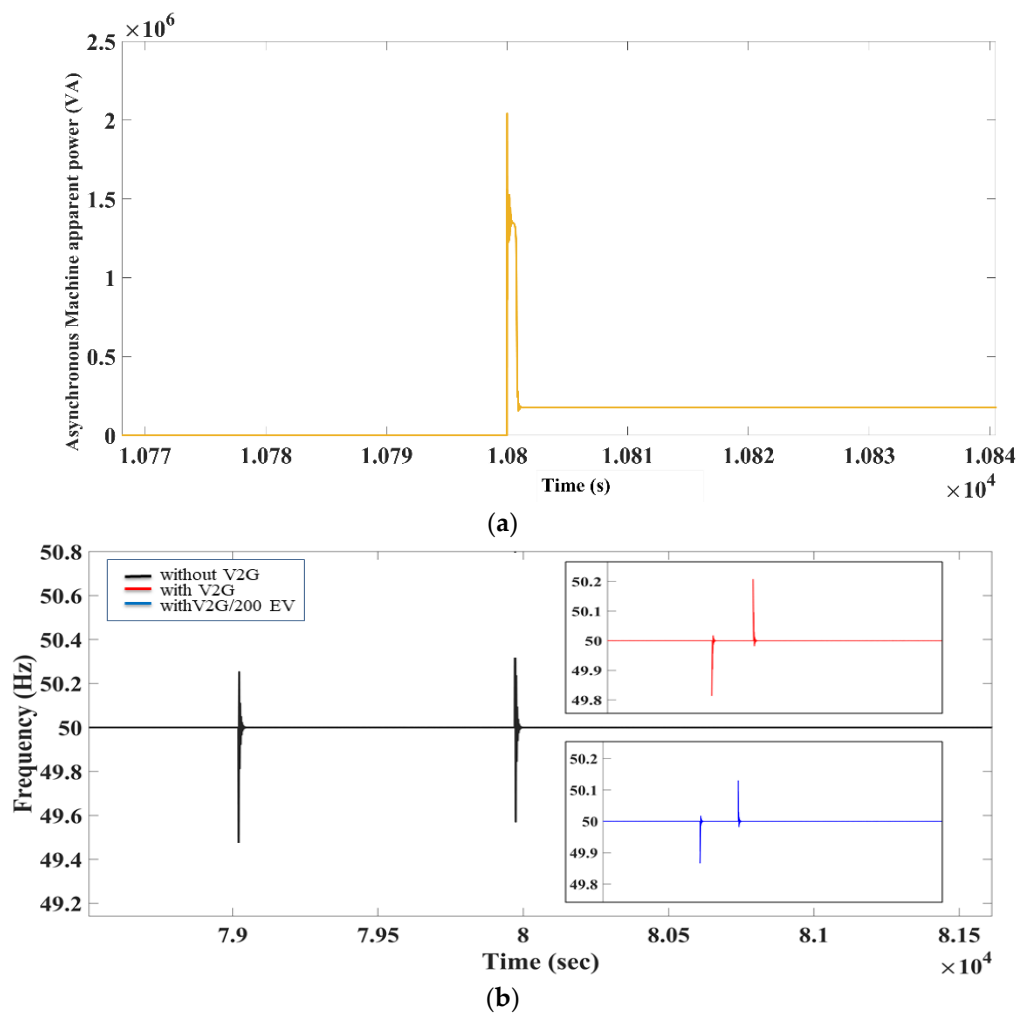

**Figure 9.** Starting of asynchronous machine. (**a**) Apparent power and (**b**) frequency regulation for three different cases.

Figure 9b depicts an industrial microgrid frequency in the absence of V2G mode. When V2G was activated, the EVs begin to contribute to the regulation of grid and reduce frequency deviations to lower levels, indicating grid frequency stabilization, as illustrated in Figure 9a. Noticeable improvements in frequency deviations are shown in Figure 9b. As in the preceding scenario, raising the number of EVs in fleets reduces frequency deviations even further. When additional 100 EVs are added to the fleet, the frequency variations improve even more, from 49.80 Hz to the 49.86 Hz.

This illustrates that by adding more cars, a frequency deviation could be lowered even further to the more minimal value. Table 6 presents a summary for the deviations of frequency for all the cases. The result obtained by proposed controller has better frequency regulation compared with reference method. This Figure 9 clearly shows how an increase in the number of EVs in the fleet might result in significant grid improvements.

**Table 6.** Comparison of results for three separate cases in scenario 3 using the proposed controller compared to those of Reference.

| Mode | V2G Off | | V2G On (100 EVs) | | V2G On (200 EVs) | |
|---|---|---|---|---|---|---|
| | Ref. [51] | Proposed Controller | Ref. [51] | Proposed Controller | Ref. [51] | Proposed Controller |
| Frequency (Min Value) | 49.40 | 49.44 | 49.77 | 49.80 | 49.83 | 49.86 |
| Frequency | - | 50.28 | - | 50.20 | - | 50.14 |

(Max Value)

## 8. Discussion

The above section of this research article focuses on the various situations that were simulated over the course of a day. The proposed primary frequency controller using FOPID has been checked for frequency regulation. Three different scenarios and cases are used to validate the above work. This scenario has the following elements: (1) PV Farm power output has decreased, (2) wind turbine tripping, and (3) starting of the asynchronous load. Each scenario in the microgrid is tested for frequency regulation in the three separate cases. V2G mode is disabled in the first case. In the second situation, V2G mode is enabled with 100 EVs, and in the third case, the number of EV's in V2G mode is raised (200 EV's). The simulation results show that the proposed controller can effectively support an industrial microgrid in terms of frequency regulation in three different scenarios. Furthermore, simulation results show that increasing the number of electric vehicles in the fleets during V2G mode can boost the frequency of an industrial microgrid to even higher levels. The suggested primary frequency controller with FOPID gives a better result than the other frequency regulation methods discussed in the introduction section. For comparison, the findings of this study are compared to the most recent work reference [51]. This work is simulated, and results are analyzed using MATLAB/SIMULINK.

## 9. Conclusions

EVs are integrated into the microgrid for proposed primary frequency regulation in this research article. Microgrid frequency deviates due to the intermittency of solar and wind energy systems. For primary frequency regulation, a proposed controller with V2G technology is used. V2G with the proposed controller can eliminate frequency fluctuation and regulate system frequency because of its quick response. The proposed primary frequency controller using FOPID is used to accomplish this project. An industrial microgrid is considered, including wind farms, PV farms, diesel generator, loads, and storage systems. The impact of various contingencies on a primary frequency has been simulated and observed using three different scenarios and cases. According to simulation results, when the V2G approach is activated for EV charging or discharging, the frequency was well regulated inside the permissible margin, compared with when a V2G approach is disabled. By raising the number of electric vehicles in the network, frequency regulation improves even more, since more vehicles contribute towards grid regulation capability.

**Author Contributions:** Conceptualization, S.S.S., V.K.J., J.M.P.-P. and F.P.G.M.; methodology, S.S.S. and V.K.J., H.M.; software, S.S.S. and V.K.J.; validation, J.N.S., S.S.S. and V.K.J.; formal analysis, S.S.S. and V.K.J. , H.M.; investigation, S.S.S. and V.K.J.; resources, S.S.S. and V.K.J., H.M., J.M.P.-P. and F.P.G.M.; data curation, S.S.S., V.K.J., H.M., J.M.P.-P. and F.P.G.M.; writing—original draft preparation, S.S.S., V.K.J., J.M.P.-P., H.M. and F.P.G.M.; writing—review and editing, J.N.S., V.K.J., J.M.P.-P., H.M. and F.P.G.M.; visualization, J.N.S., S.S.S., H.M. and V.K.J.; supervision, J.N.S. and V.K.J.; project administration, J.M.P.-P. H.M. and F.P.G.M.; funding acquisition, J.M.P.-P. and F.P.G.M. All authors have read and agreed to the published version of the manuscript.

**Funding:** The work reported herein has been financially supported by DIRECCIÓN GENERAL DE UNIVERSIDADES, INVESTIGACIÓN E INNOVACIÓN OF Castilla-La Mancha, under Research Grant ProSeaWind project (Ref.: SBPLY/19/180501/000102).

**Institutional Review Board Statement:** Not applicable.

**Informed Consent Statement:** Not applicable.

**Data Availability Statement:** Not applicable.

**Acknowledgments:** The authors would like to acknowledge the support from Intelligent Prognostic Private Limited India Researcher's Supporting Project.

**Conflicts of Interest:** The authors declare no conflict of interest.

**Abbreviations**

| | |
|---|---|
| ACM | Asynchronous Machine |
| CTC | Compare to Constant |
| DER | Distributed Energy Resources |
| DG | Diesel Generator |
| ESS | Energy Storage System |
| EV | Electric Vehicle |
| G2V | Grid to Vehicle |
| IMGs | Industrial Microgrid |
| MG | Microgrid |
| PCC | Point of Common Coupling |
| PFC | Primary Frequency Control |
| PV | Photovoltaic Panel |
| RES | Renewable Energy Sources |
| SOC | State of Charge |
| V2G | Vehicle-to-Grid |